\documentclass[letterpaper, 10 pt, conference]{ieeeconf} 
\IEEEoverridecommandlockouts                         
\usepackage{times}  
\usepackage{helvet}  
\usepackage{courier}  
\usepackage[hyphens]{url}  
\usepackage[hidelinks]{hyperref}
\usepackage{graphicx} 
\usepackage{subcaption}
\usepackage{multirow}
\usepackage{multicol}
\usepackage{algorithm}
\usepackage{algorithmic}
\usepackage{amsmath}
\usepackage{amsfonts}
\usepackage{xcolor}

\usepackage{newfloat}
\usepackage{listings}
\usepackage{cite}
\usepackage{booktabs}
\usepackage{makecell}
\newtheorem{theorem}{Theorem}

\newtheorem{lemma}{Lemma}

\newtheorem{problem}{Problem}
\newtheorem{example}{Example}

\newcommand{\relu}{\text{ReLU}}
\title{Learning Neural Network Barrier Functions with Termination Guarantees}

\title{\LARGE \bf
Verification-Aided Learning  of Neural Network Barrier Functions with \\
Termination Guarantees
}

\author{Shaoru Chen$^{1}$, Lekan Molu$^{1}$, Mahyar Fazlyab$^{2}$
\thanks{$^{1}$ Shaoru Chen and Lekan Molu are with Microsoft Research, 300 Lafayette Street, New
York, NY, 10012, USA. Email: {\tt\small \{shaoruchen, lekanmolu\}@microsoft.com}.}%
\thanks{$^{2}$ Mahyar Fazlyab is with the Mathematical Institute for Data Science, Johns Hopkins University, USA. Email: {\tt\small mahyarfazlyab@jhu.edu}.}%
}

\begin{document}

\maketitle

\begin{abstract}
Barrier functions are a general framework for establishing a safety guarantee for a system. However, there is no general method for finding these functions. To address this shortcoming, recent approaches use self-supervised learning techniques to learn these functions using training data that are periodically generated by a verification procedure, leading to a verification-aided learning framework. Despite its immense potential in automating barrier function synthesis, the verification-aided learning framework does not have termination guarantees and may suffer from a low success rate of finding a valid barrier function in practice. In this paper, we propose a holistic approach to address these drawbacks. With a convex formulation of the barrier function synthesis, we propose to first learn an empirically well-behaved NN basis function and then apply a fine-tuning algorithm that exploits the convexity and counterexamples from the verification failure to find a valid barrier function with \emph{finite-step termination guarantees}: if there exist valid barrier functions, the fine-tuning algorithm is guaranteed to find one in a finite number of iterations. We demonstrate that our fine-tuning method can significantly boost the performance of the verification-aided learning framework on examples of different scales and using various neural network verifiers.

\end{abstract}

\section{Introduction}
\label{sec:introduction}

\begin{figure}[tb]
    \centering
    \includegraphics[width=\columnwidth]{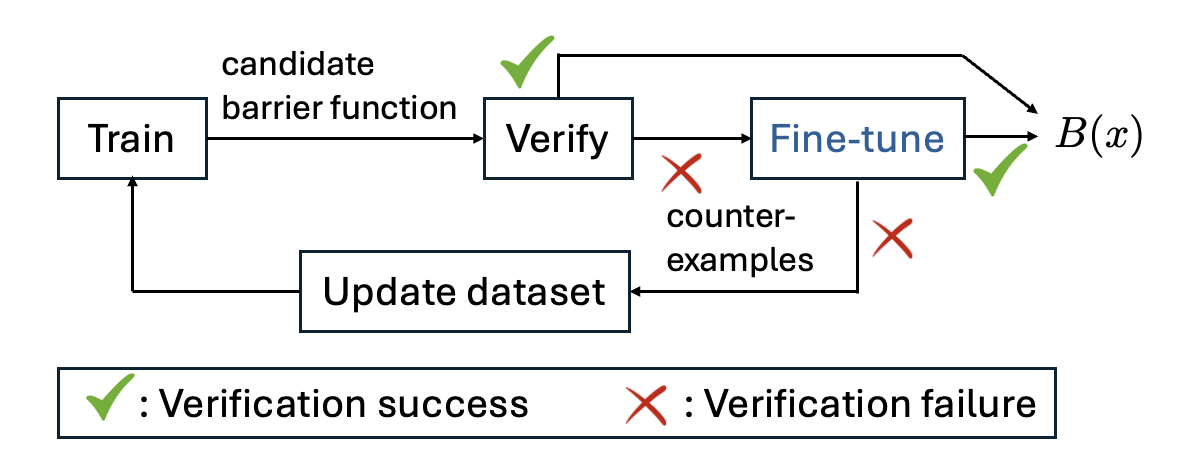}
    \caption{The verification-aided learning framework with our proposed fine-tuning method (highlighted in blue). }
    \label{fig:framework}
\end{figure}

Providing safety and stability guarantees is a central goal in controller design for dynamical systems. These guarantees are often established by finding certificate functions. In particular, barrier functions aim to prove that the trajectories of an autonomous dynamical system remain inside a safe set over an infinite time horizon. Similarly, control barrier functions provide a general framework for designing control inputs that render a safe set invariant. Despite their generality, choosing an appropriate functional form and finding its parameters requires either computationally prohibitive methods or case-by-case analysis based on expert knowledge. To mitigate this, recent approaches propose to learn certificate functions from data using function approximators such as neural networks (NNs)~\cite{dawson2023safe}. However, to achieve deterministic guarantees, these approaches must be accompanied by a verification procedure, which either proves that the certificate is valid or generates counterexamples that can be added to the dataset for retraining the certificate function candidate~\cite{abate2020formal, abate2021fossil, Dai2021lyapunov, zhao2022verifying, liu2023safe}. We refer to this general framework of learning certificate functions with formal guarantees as \emph{verification-aided learning} and summarize its schematics in Fig.~\ref{fig:framework}. Despite the immense potential of this verification-aided learning framework in automating the certificate learning process, there is no guarantee that the updated certificate function will be `closer' to a valid one in each iteration. Therefore, the alternation between training and verification may never terminate.

\begin{figure}[tb]
    \centering
    \includegraphics[width=\columnwidth]{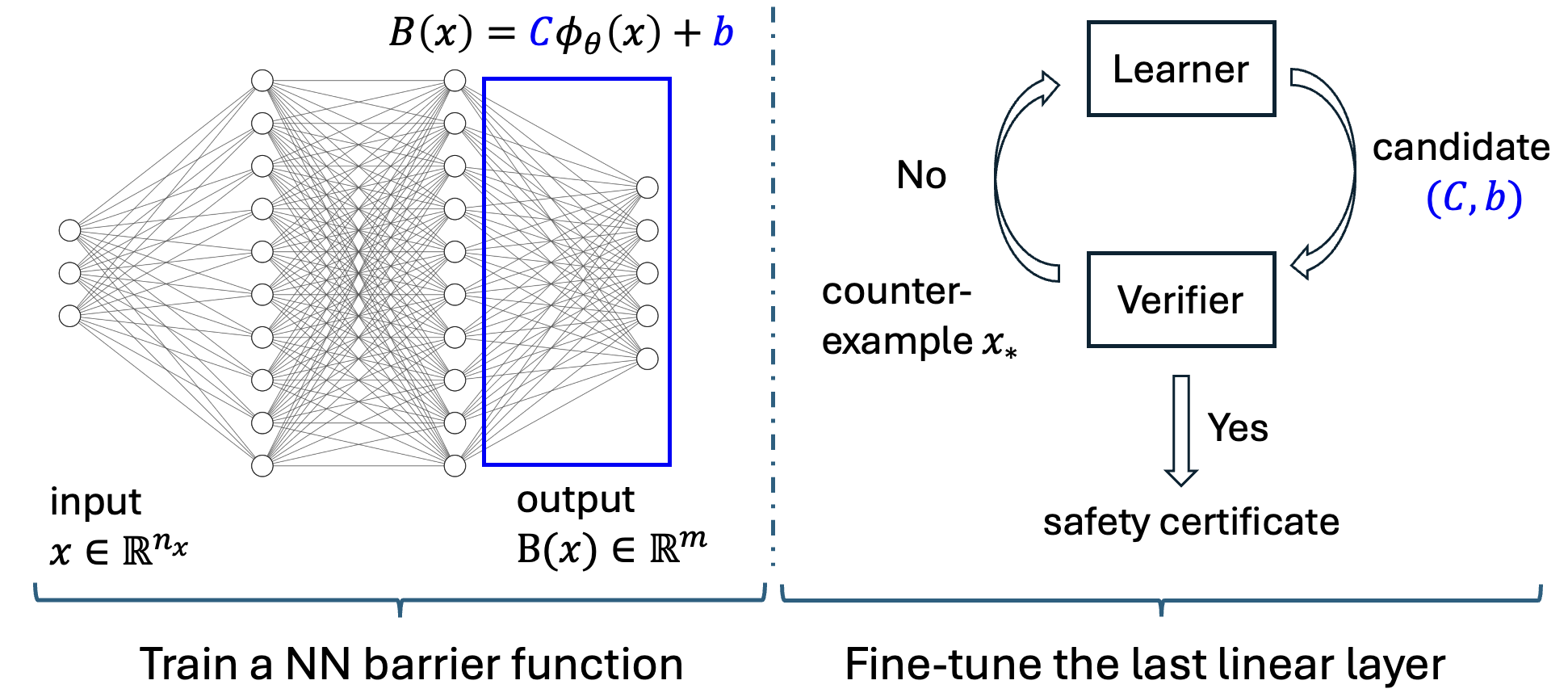}
    \caption{Illustration of the fine-tuning method. After training the NN barrier function (left), we fine-tune the last linear layer through a counterexample-guided inductive synthesis framework (right) which enjoys termination guarantees.}
    \label{fig:overview}
\end{figure}

Motivated by this challenge, in this work, we propose a principled approach to fine-tune the learned NN certificate function candidate with `monotonic' improvement guarantees. In particular, we consider safety verification of a NN policy with the goal of finding a valid barrier function to certify avoidance of unsafe sets. As illustrated in Fig.~\ref{fig:overview}, after training the NN barrier function, we propose to fine-tune the last linear layer of the NN through a counterexample-guided inductive synthesis (CEGIS) framework~\cite{solar2006combinatorial, chen2021learning} with termination guarantees, i.e., if there exist valid barrier functions, the fine-tuning algorithm is guaranteed to find one in a finite number of steps. To achieve such a guarantee, we formulate a NN \emph{vector barrier function}, inspired by the related work in safety analysis for continuous-time systems~\cite{sogokon2018vector}, which maintains the \emph{convexity} of the barrier function formulation and enables flexible tuning of the expressivity of the function class, e.g., by adjusting the architecture of the NN. The intuition behind our method in Fig.~\ref{fig:overview} is that we can empirically learn a good barrier function candidate from data and efficiently close the gap of providing formal guarantees through convex optimization. The \textbf{contributions} of this paper are summarized as follows. 
\begin{enumerate}
\item We propose a novel verification-aided learning framework with a fine-tuning step to learn NN certificate functions with formal guarantees. The proposed fine-tuning approach follows an analytic center cutting-plane method~\cite{atkinson1995cutting} and enjoys finite-step termination guarantees. This enables our framework to benefit from both the learning capacity of NNs and performance guarantees from convex optimization. 
\item We formulate a NN vector barrier function for safety analysis of discrete-time autonomous systems. This formulation allows easy tuning of the function class complexity and enables posing the barrier function synthesis problem as a convex one. 
\item We demonstrate that the proposed fine-tuning method can significantly boost the success rate and runtime of the verification-aided learning framework on numerous examples of varying scales and with different NN verifiers, namely a mixed-integer programming (MIP)-based verifier~\cite{tjeng2018evaluating} and a GPU-accelerated verifier $\alpha,\beta$-CROWN~\cite{wang2021beta}.
\end{enumerate}

\subsection{Related work}
\paragraph{Learning certificate functions} Certificate functions such as (control) barrier/Lyapunov functions
are powerful tools for certifying the safety/stability of a dynamical system or designing safe/stabilizing controllers. Motivated by the difficulty of finding certificate functions for general nonlinear systems, learning a NN certificate function has been investigated in~\cite{chang2019neural, jin2020neural, robey2020learning, qin2020learning, lindemann2021learning, gaby2022lyapunov, zhang2023compositional, xiao2023barriernet} with promising results. However, in these works, verification of the learned certificate functions is treated as a separate task, which brings the question of refining the learned NN once the verification fails. 

\paragraph{Verification of NN certificate functions}
NN verification concerns formally certifying the properties of a trained NN~\cite{liu2021algorithms}. The rapid development of NN verification in recent years has provided us with a rich set of tools with constantly improving scalability and numerical efficiency~\cite{xu2020automatic, wang2021beta, zhang2022general}. 
Going beyond verification, in this work, we consider synthesizing a NN certificate function using the NN verification tools as a sub-routine.

\paragraph{Counterexample-guided synthesis}
Counterexample-guided inductive synthesis (CEGIS)~\cite{solar2006combinatorial} is a general framework that utilizes the information or counterexamples from verification failure to progressively refine the solutions. 
CEGIS has been widely applied to NN certificate function synthesis in~\cite{dai2020counter, Dai2021lyapunov, abate2020formal, abate2021fossil, peruffo2021automated, liu2023safe}. Although the counterexamples are believed to be informative and useful for the learner, there is often no guarantee that CEGIS will find a feasible solution even if there exists one. From the theoretical perspective, \cite{chen2021learning, ravanbakhsh2019learning, chen2021learningROA} show that such a guarantee can be achieved following a cutting-plane method~\cite{atkinson1995cutting} with a convex parameterization of the certificate function, but they do not consider how such results can boost learning of a NN certificate function. 



\section{Problem Statement}
We consider the dynamical system 
\begin{equation} \label{eq:nonlinear_sys}
x_{k+1} = f(x_k, u_k)
\end{equation}
where $x_k \subseteq \mathbb{R}^{n_x}$ is the state and $u_k \in \mathbb{R}^{n_u}$ is the control input. We assume a policy $u_k = \pi(x_k)$ is given, which gives rise to the autonomous, closed-loop dynamics 
\begin{equation} \label{eq:cl_dynamics}
    x_{k+1} = f_\pi(x_k) := f(x_k, \pi(x_k)).
\end{equation}
Let $\mathcal{X} \subseteq \mathbb{R}^{n_x}$ be a compact set denoting the workspace of the system, and $\mathcal{X}_0, \mathcal{X}_u \subset \mathcal{X}$ denote the set of initial conditions and unsafe sets, respectively. We call system~\eqref{eq:cl_dynamics} \emph{safe} if all trajectories in the workspace starting from the initial set $\mathcal{X}_0$ will never reach the unsafe set $\mathcal{X}_u$. 


\begin{problem}[Safety Verification] \label{prob:safety}
Given system~\eqref{eq:cl_dynamics}, the workspace $\mathcal{X}$, the initial set $\mathcal{X}_0 \subseteq \mathcal{X}$ and the unsafe set $\mathcal{X}_u \subseteq \mathcal{X}$, show that the system is safe, i.e., if $x_0 \in \mathcal{X}_0$, for all $k \geq 0$, we have $x_k \in \mathcal{X} \Rightarrow x_k \notin \mathcal{X}_u$, which means the state does not enter the unsafe region as long as it stays inside $\mathcal{X}$.
\end{problem}

Verifying the safety of the policy $\pi(\cdot)$ requires analyzing the behavior of system~\eqref{eq:cl_dynamics} over an infinite horizon. Following~\cite{prajna2004safety}, we aim to find a barrier function as a certificate of safety. However, the nonlinearity of the closed-loop dynamics brings significant challenges in numerical computation. While a NN barrier function can be parameterized and learned using empirical risk minimization (ERM), a formal verification of the trained NN barrier function is required. In case of verification failure, the whole NN barrier function must be retrained and such a `train-then-verify' procedure may never terminate. 

In this work, we propose a more principled way to learn a NN barrier function with improved efficiency. In Section~\ref{sec:vector_barrier}, we formulate and train a NN vector barrier function, and propose a fine-tuning method with termination guarantees in Section~\ref{sec:fine_tuning}. In Section~\ref{sec:training_framework}, we present the proposed verification-aided learning framework and implementation details. Numerical examples are provided in Section~\ref{sec:simulation}. Section~\ref{sec:conclusion} concludes the paper.


\section{NN Vector Barrier Function}
\label{sec:vector_barrier}
First introduced in~\cite{prajna2004safety}, the barrier function has been a popular method for safety verification of autonomous systems. Although rich theories and applications have been explored for continuous-time systems, barrier functions for discrete-time systems have received less attention. In this section, we introduce formulations of discrete-time barrier functions and propose a discrete-time vector barrier function adapted from~\cite{sogokon2018vector} to maintain the convexity of the formulation. 

\subsection{Scalar barrier functions} 
\label{sec:scalar_barrier}
Certifying the safety of system~\eqref{eq:cl_dynamics} requires showing all trajectories $\{x_k\}_{k=0}^\infty$ starting from $\mathcal{X}_0$ never reach the unsafe set $\mathcal{X}_u$. Since the infinite horizon trajectory is considered, explicitly evolving the dynamics for verification is intractable. Instead, we look for a numerical certificate that is sufficient for safety verification. 


\begin{theorem}[Scalar barrier function] \label{thm:scalar_barrier}
Let the system~\eqref{eq:nonlinear_sys} and sets $\mathcal{X}, \mathcal{X}_u, \mathcal{X}_0$ be given. If there is a continuous scalar function $B(x): \mathcal{X} \mapsto \mathbb{R}$ satisfying
	\begin{subequations} \label{eq:barrier_cond}
		\begin{align}
		\begin{split}\label{eq:barrier_cond_1}
		B(x) > 0, \forall x \in \mathcal{X}_u,
		\end{split} \\
		\begin{split}\label{eq:barrier_cond_2}
		B(x) \leq 0, \forall x \in \mathcal{X}_0,
		\end{split} \\
		\begin{split}\label{eq:barrier_cond_3}
		B(f_\pi(x)) \leq  \gamma B(x), \forall x \in \mathcal{X},
		\end{split}
		\end{align}
	\end{subequations}
    where $\gamma \geq 0$, then the safety of system~\eqref{eq:cl_dynamics} is guaranteed. Such a $B(x)$ satisfying conditions~\eqref{eq:barrier_cond_1} to~\eqref{eq:barrier_cond_3} is called a barrier function. 
\end{theorem}

\begin{proof}
For the barrier function $B(x)$, we have $x_0 \in \mathcal{X}_0$ and $B(x_0) \leq 0$ by condition~\eqref{eq:barrier_cond_2}. From condition~\eqref{eq:barrier_cond_3}, we have that for all $k > 0$ and $x_0, x_1, \cdots, x_{k-1} \in \mathcal{X}$, $B(x_k) \leq \gamma^k B(x_0) \leq 0$ which indicates that $x_k \notin \mathcal{X}_u$. 
\end{proof}

\paragraph{Convex formulation of scalar barrier function} 


From the definition of barrier function in Theorem~\ref{thm:scalar_barrier}, it follows that the set of barrier functions is \emph{convex}, i.e., $\lambda B_1(x) + (1 - \lambda) B_2(x)$ with $0 \leq \lambda \leq 1$ is a barrier function if both $B_1(x)$ and $B_2(x)$ satisfy the conditions~\eqref{eq:barrier_cond_1} to~\eqref{eq:barrier_cond_3}. This convexity property facilitates the design of numerically efficient methods to search $B(x)$. For example, the continuous-time counterparts of constraints~\eqref{eq:barrier_cond} enable finding $B(x)$ through semidefinite programming for polynomial dynamical systems~\cite{prajna2004safety, prajna2007framework, kong2013exponential}. One important feature of the convex barrier function formulation is that if we parameterize a candidate barrier function as $B(x) = \sum_{i=1}^M c_i \phi_i(x)$, where $c = [c_1 \ \cdots \ c_M]^\top \in \mathbb{R}^M$ denotes the coefficient vector and $\phi_i(x): \mathbb{R}^{n_x} \mapsto \mathbb{R}$ denotes a nonlinear basis function, the set of valid barrier functions in the parameter space of $c$ is also convex. In this work, we will consider parameterizing $B(x)$ as a NN~\footnote{The bias term in~\eqref{eq:scalar_NN} can be interpreted as the coefficient of the basis function $\phi_M(x)= 1$.}, i.e., 
\begin{equation} \label{eq:scalar_NN}
    B(x) = c^\top \phi_\theta(x) + b, 
\end{equation}
where $\phi_\theta(x):\mathbb{R}^{n_x} \mapsto \mathbb{R}^M$ is an MLP parameterized by $\theta$ and $c, b$ can be interpreted as the weights and bias of the last linear layer. 

\paragraph{Effects of the parameter $\lambda$} 
Despite the simple form of the barrier function conditions~\eqref{eq:barrier_cond}, the choice of $\gamma$ in~\eqref{eq:barrier_cond_3} has a non-trivial impact on the barrier function synthesis. In parallel to the observations made in~\cite{kong2013exponential} on continuous-time systems, we discuss the implications of different values of $\lambda$ for the discrete-time autonomous dynamics~\eqref{eq:cl_dynamics} below. 

\begin{enumerate}
    \item Case $\gamma = 0$: Denote $\Omega = \{x \vert B(x) \leq 0 \cap \mathcal{X}\}$ the sublevel set of a barrier function $B(x)$ and $\Omega^\mathsf{c} = \mathcal{X} \setminus \Omega$ its complement. When $\gamma = 0$, \eqref{eq:barrier_cond_3} requires that $B(x_1) \leq 0$ for all $x_0 \in \mathcal{X}$, i.e., the system~\eqref{eq:cl_dynamics} evolves into $\Omega$ in one step starting from the state space $\mathcal{X}$, and this condition tends to be overly conservative.
    \item Case $0 < \gamma < 1$: With a positive scalar $\gamma$ and a trajectory $\{x_k\}_{k=0}^\infty$ contained in $\mathcal{X}$, condition~\eqref{eq:barrier_cond_3} indicates that $B(x_k) \leq \gamma^k B(x_0)$ for all $x_0 \in \mathcal{X}$. It follows that $\lim_{k\rightarrow \infty} B(x_k) \leq 0$ and the sublevel set $\Omega$ of $B(x)$ is attractive for all $x_0 \in \mathcal{X}$.
    \item Case $\gamma > 1$: The upper bound $\gamma^k B(x_0)$ diverges to $-\infty$ for $x_0 \in \Omega$, which means all trajectories starting inside $\Omega$ have to leave $\mathcal{X}$ in a finite number of steps and $\Omega$ cannot contain an equilibrium. 
\end{enumerate}

It follows that for the set of barrier functions to be non-empty, the choice of $\gamma$ should depend on the dynamics of the autonomous system~\eqref{eq:cl_dynamics}. Meanwhile, with the scalar barrier function formulation, the flexibility of tuning the hyperparameter $\gamma$ is limited. This leads to the conservatism of the scalar barrier function formulation. 


In the next subsection, we introduce vector barrier functions for safety verification which allow us to search for safety certificates with less conservatism while maintaining the convexity of the barrier function formulation. 

\subsection{Vector barrier functions}
A vector barrier function $B(x): \mathbb{R}^{n_x} \rightarrow \mathbb{R}^{m}$ is defined as $B(x) = [B_1(x) \ B_2(x) \ \cdots \ B_m(x)]^\top$ where $B_1(x), \cdots, B_m(x)$ are scalar functions mapping from $\mathbb{R}^{n_x}$ to $\mathbb{R}$. As an adaption of the vector barrier function for continuous-time systems~\cite{sogokon2018vector}, we define the discrete-time vector barrier function in the following lemma.

\begin{lemma}
Let the system~\eqref{eq:nonlinear_sys} and sets $\mathcal{X}, \mathcal{X}_u, \mathcal{X}_0$ be given. A vector-valued function $B(x) = [B_1(x) \ B_2(x) \ \cdots \ B_m(x)]^\top: \mathcal{D} \mapsto \mathbb{R}^{m}$ is called a vector barrier function if it satisfies
\begin{subequations} \label{eq:vec_barrier_cond}
	\begin{align}
	\begin{split}\label{eq:vec_barrier_cond_1}
	 \vee_{i = 1}^m B_i(x) > 0, \forall x \in \mathcal{X}_u,
	\end{split} \\
	\begin{split}\label{eq:vec_barrier_cond_2}
	\wedge_{i=1}^m B_i(x) \leq 0, \forall x \in \mathcal{X}_0, 
	\end{split} \\
	\begin{split}\label{eq:vec_barrier_cond_3}
	B(x_+) \leq A B(x), \forall x \in \mathcal{X},
	\end{split}
	\end{align}
\end{subequations}
where $A \in \mathbb{R}^{m \times m}$ is a non-negative matrix, i.e., $A \geq 0$, $\vee$ denotes the disjunction operator meaning $\vee_{i = 1}^m B_i(x) > 0 \Leftrightarrow \exists \ i^* \in \{1, \cdots, m\} \text{ s.t. } B_{i^*} > 0$, and $\wedge$ denotes the conjunction operator meaning $\wedge_{i=1}^m B_i(x) \leq 0 \Leftrightarrow B_i(x) \leq 0, i = 1, \cdots,m$. Then the existence of a vector barrier function $B(x)$ certifies the safety of the system. 
\end{lemma}

\begin{proof}
Condition~\eqref{eq:vec_barrier_cond_2} indicates that $B(x_0) \leq 0$. Since $A$ is a non-negative matrix, we have $B(x_k) \leq A^k B(x_0) \leq 0$ for all $k \geq 0$ as long as $x_k \in \mathcal{X}$. It follows that $x_k$ cannot reach $\mathcal{X}_u$ where at least one entry of $B(x_k)$ is positive.
\end{proof}

It is obvious that the vector barrier function contains the scalar barrier function defined in~\eqref{eq:barrier_cond} as a subclass $m = 1$. However, when $m > 1$, the set of vector barrier functions is not convex due to the disjunction operator in~\eqref{eq:vec_barrier_cond_1}. To recover convexity, we apply the following formulation with stronger conditions:
\begin{theorem}
The system~\eqref{eq:nonlinear_sys} with the workspace $\mathcal{X}$, the initial set $\mathcal{X}_0$ and the unsafe set $\mathcal{X}_u$ is safe if there exists a vector-value function $B(x):\mathbb{R}^{n_x}\mapsto \mathbb{R}^{m}$ satisfying 
\begin{subequations} \label{eq:cvx_vec_barrier_cond}
	\begin{align}
	\begin{split}\label{eq:cvx_vec_barrier_cond_1}
	 B_{i^*}(x) > 0, \forall x \in \mathcal{X}_u, 
	\end{split} \\
	\begin{split}\label{eq:cvx_vec_barrier_cond_2}
	\wedge_{i=1}^m B_i(x) \leq 0, \forall x \in \mathcal{X}_0,
	\end{split} \\
	\begin{split}\label{eq:cvx_vec_barrier_cond_3}
	B(x_+) \leq A B(x), \forall x \in \mathcal{X},
	\end{split}
	\end{align}
\end{subequations}
where $i^* \in \{1, \cdots, m\}$ is a given number and $A$ is a non-negative matrix.
\end{theorem}

In the rest of the paper, a vector barrier function is referred to as any $B(x)$ that satisfies conditions~\eqref{eq:cvx_vec_barrier_cond}. It can be easily verified that the set of valid $B(x)$ satisfying~\eqref{eq:cvx_vec_barrier_cond} is convex. In addition to tuning the function class of each scalar barrier function $B_i(x)$, we can systematically increase the expressivity of the vector barrier function by increasing the dimension $m$ as well. The benefit of the vector barrier function is illustrated in the following example, where the safety of a linear system can be certified by a vector $B(x)$ using only linear functions while a scalar barrier function is required to be at least quadratic. 

\begin{figure}
\begin{subfigure}{0.49 \columnwidth}
    \centering
    \includegraphics[width=\textwidth]{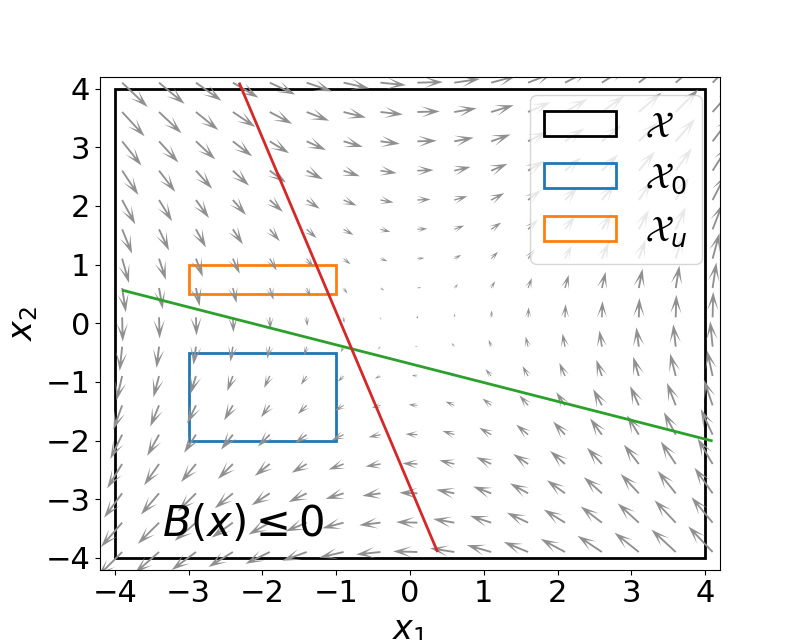}
    \caption{Vector barrier function.}
    \label{fig:linear_system_vbf}
\end{subfigure}
\hfill
\begin{subfigure}{0.49 \columnwidth}
    \centering
    \includegraphics[width=\textwidth]{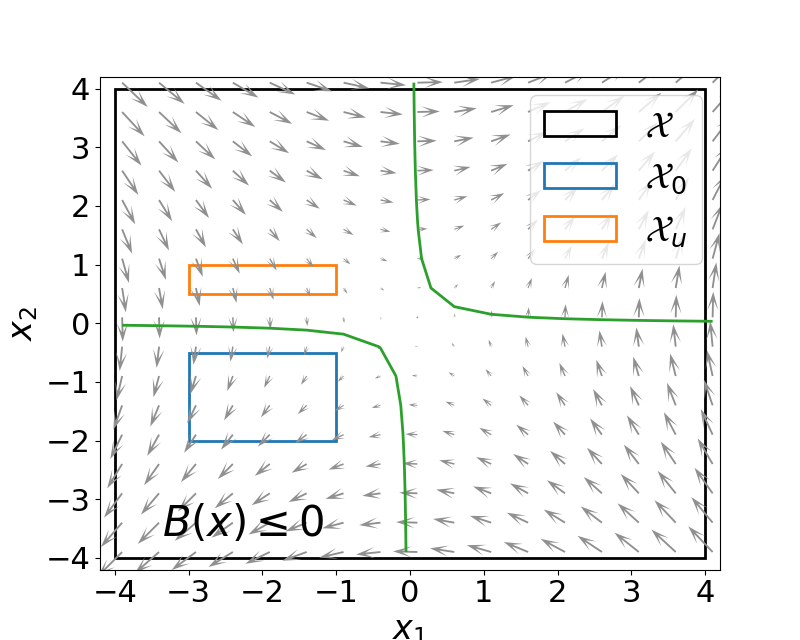}
    \caption{Scalar barrier function.}
    \label{fig:linear_system_sbf}
\end{subfigure}
\caption{The $0$-level sets of the \emph{vector} barrier function using linear functions (left) and \emph{scalar} barrier function using a quadratic function (right) for safety certification of the system considered in Example~\ref{example:linear} are plotted. The lower-left, encircled regions given by $\{x \mid B(x) \leq 0\}$ separate the trajectories starting from the initial set from the unsafe set.  }
\label{fig:linear_system}
\end{figure}

\begin{example} \label{example:linear}
Consider a linear system $x_{k+1} = \bigl[ \begin{smallmatrix}1 & 1\\ 0 & 1\end{smallmatrix}\bigr ] x_k$ with the workspace $\mathcal{X}$, initial set $\mathcal{X}_0$, and unsafe set $\mathcal{X}_u$ shown in Fig.~\ref{fig:linear_system}. There is no linear scalar barrier function satisfying~\eqref{eq:barrier_cond} that can certify the safety of the system, while a vector barrier function with dimension $2$ using two linear functions $B_1(x), B_2(x)$ and $A = \bigl[ \begin{smallmatrix}1 & 1\\ 0 & 1\end{smallmatrix}\bigr ]$ succeeds (Fig.~\ref{fig:linear_system_vbf}). To find a scalar barrier function, a quadratic function parameterization is needed as shown in Fig.~\ref{fig:linear_system_sbf}.
\end{example}

\subsection{Training of NN vector barrier functions}
We now parameterize a NN vector barrier function as 
\begin{equation} \label{eq:nn_vbf}
    B(x) = C \phi_\theta(x) + b,
\end{equation}
where $\phi_\theta(x):\mathbb{R}^{n_x} \mapsto \mathbb{R}^M$ is an MLP whose weights and biases are denoted by the parameter $\theta$, and $C \in \mathbb{R}^{m \times M}, b \in \mathbb{R}^m$ are the weights and bias of the last linear layer, respectively. Here we highlight the parameters of the last linear layer which we will fine-tune in the next section. As highlighted in the discussion of the scalar barrier function, the hyperparameter $A$ in~\eqref{eq:cvx_vec_barrier_cond_3} has a complex dependence on the underlying dynamics~\eqref{eq:cl_dynamics} and a feasible value is generally difficult to find. We propose to train the NN barrier function~\eqref{eq:nn_vbf} jointly with the hyperparameter $A$ using the samples collected from the workspace $\mathcal{X}$. Without loss of generality, we fix $i^* = 1$ in constraint~\eqref{eq:cvx_vec_barrier_cond_1}.

We consider the following empirical barrier function risk
\begin{equation} \label{eq:erm}
\begin{aligned}
L(\theta, C, b, A) &= \frac{1}{N_0} \sum_{i=1}^{N_0} \mathbf{1}^\top \relu (B(x_0^i)) + \\
& \frac{1}{N_u} \sum_{i=1}^{N_u} e_{1}^\top \relu(-B(x_u^i)) + \\
& \frac{1}{N}\sum_{i=1}^N \mathbf{1}^\top \relu(B(f_\pi(x^i))-AB(x^i)),
\end{aligned}
\end{equation}
where $\mathbf{1}$ denotes the vector of all ones, $e_1 = [1 \ 0 \ \cdots \ 0]^\top$ denotes the first standard basis, $\{x_0^i\}_{i=1}^{N_0}$, $\{ x_u^i\}_{i=1}^{N_u}$, $\{ x^i\}_{i=1}^{N}$ are the $N_0$ samples from the initial set $\mathcal{X}_0$, $N_u$ samples from the unsafe set $\mathcal{X}_u$, and $N$ samples from the workspace $\mathcal{X}$, respectively. Then, projected gradient descent (PGD) is applied to minimize $L(\theta, C, b, A)$ subject to $A \geq 0$ to generate a NN vector barrier function candidate $B(x)$. In the next section, we will address the problem of verifying the validity of the trained NN $B(x)$ and updating $B(x)$ in case of verification failure. 


\section{Fine-Tuning through Convex Optimization}
\label{sec:fine_tuning}
With the trained NN barrier function candidate $B(x)$ and the parameter $A$, we can verify if $B(x)$ satisfies all the constraints in~\eqref{eq:cvx_vec_barrier_cond} using NN verification tools. However, in case of verification failure, retraining of $B(x)$ is required and this `train-then-verify' process may never terminate with a valid barrier function. To address this issue, in this section, we propose a CEGIS method to only update the last linear layer $(C, b)$ for the following reasons:
\begin{enumerate}
    \item Empirically, it is reasonable to hypothesize that, after sufficient training, the NN vector barrier function candidate $B(x)$ is close to being valid and only minor adjustment is needed. 
    \item Theoretically, if there exists a feasible pair $(C, b)$ such that $B(x)$ is valid, our CEGIS algorithm is guaranteed to find a feasible pair in a finite number of iterations. 
\end{enumerate}
Our algorithm of fine-tuning the last linear layer consists of a learner, which proposes a candidate $(C, b)$, and a verifier, which verifies the validity of the candidate $(C, b)$ or falsifies it with a counterexample state. Similar to~\cite{chen2021learning}, by designing the interaction between the learner and verifier according to the cutting-plane method from convex optimization, we achieve the finite-step termination guarantee. We elaborate on the algorithm design in the following subsections. 

\subsection{Feasible set of the last layer}
After training the NN vector barrier function, we fix the parameter $A$ and the subnetwork $\phi_\theta(x)$ as the basis function. For notational simplicity, we denote $W = \text{vec}(C, b)$ as the vectorized parameter of the last linear layer of $B(x)$. Our goal is to find a $W$ such that $B(x)$ satisfies all the constraints in~\eqref{eq:cvx_vec_barrier_cond}. We denote the set of feasible parameters of $W$ as 
\begin{equation}
\mathcal{F} = \{ W \in \mathbb{R}^{m \times (M+1)} \mid B(x) \text{ satisfies } \eqref{eq:cvx_vec_barrier_cond}.\}
\end{equation}
and call $\mathcal{F}$ the \emph{target set}. We observe that when the state is fixed, constraints \eqref{eq:cvx_vec_barrier_cond} are all linear constraints in $W$. Therefore, the target set $\mathcal{F}$ is described by an (uncountably) infinite number of linear constraints and is convex. Our goal is to either find a feasible point in $\mathcal{F}$, which gives a valid barrier function and proves the safety of the system~\eqref{eq:cl_dynamics}, or certify that $\mathcal{F}$ is empty, which means retraining of the basis function $\phi_\theta(x)$ or the parameter $A$ is needed. 

Although being convex, the target set $\mathcal{F}$ is defined by infinitely many linear constraints which pose numerical challenges. This motivates a sampling-based method to over-approximate $\mathcal{F}$ by considering the barrier function conditions~\eqref{eq:cvx_vec_barrier_cond} only on a set of finite samples $\mathcal{S} = \{\mathcal{S}_0, \mathcal{S}_u,  \mathcal{S}_\mathcal{X}\}$ where $\mathcal{S}_0, \mathcal{S}_u, \mathcal{S}_\mathcal{X}$ are collections of a finite number of states in $\mathcal{X}_0, \mathcal{X}_u, \mathcal{X}$, respectively. With the sample set $\mathcal{S}$, an over-approximation of the target set is given by
\begin{equation} \label{eq:localization}
\begin{aligned}
\widetilde{\mathcal{F}} &= \{W \vert B_{i^*}(x) >0 \ \forall x \in \mathcal{S}_u, \wedge_{i=1}^m B_i(x) \leq 0 \ \forall x \in \mathcal{S}_0,\\
& B(f_\pi(x)) \leq A B(x) \ \forall x \in \mathcal{S}_\mathcal{X} \}.
\end{aligned}
\end{equation}

Obviously, $\widetilde{\mathcal{F}}$ is an over-approximation of $\mathcal{F}$, and finding a feasible point in $\widetilde{\mathcal{F}}$ reduces to solving a tractable convex feasibility problem. For a given set of samples $\mathcal{S}$, if $\widetilde{\mathcal{F}}$ is shown to be empty, we certify that the target set is empty as well, i.e., $\mathcal{F} =\emptyset$. However, if $\widetilde{\mathcal{F}}$ is non-empty, we have no guarantee that any feasible point $W \in \widetilde{\mathcal{F}}$ will generate a valid barrier function even as we increase the size of the sample set $\mathcal{S}$ asymptotically to infinity. 

To fix this issue, we propose a sampling strategy that consists of a learner and a verifier based on cutting planes with which we can expand the sample set $\mathcal{S}$ iteratively.

\begin{figure}[tb]
\begin{subfigure}{0.49 \columnwidth}
    \centering
    \includegraphics[width=\textwidth]{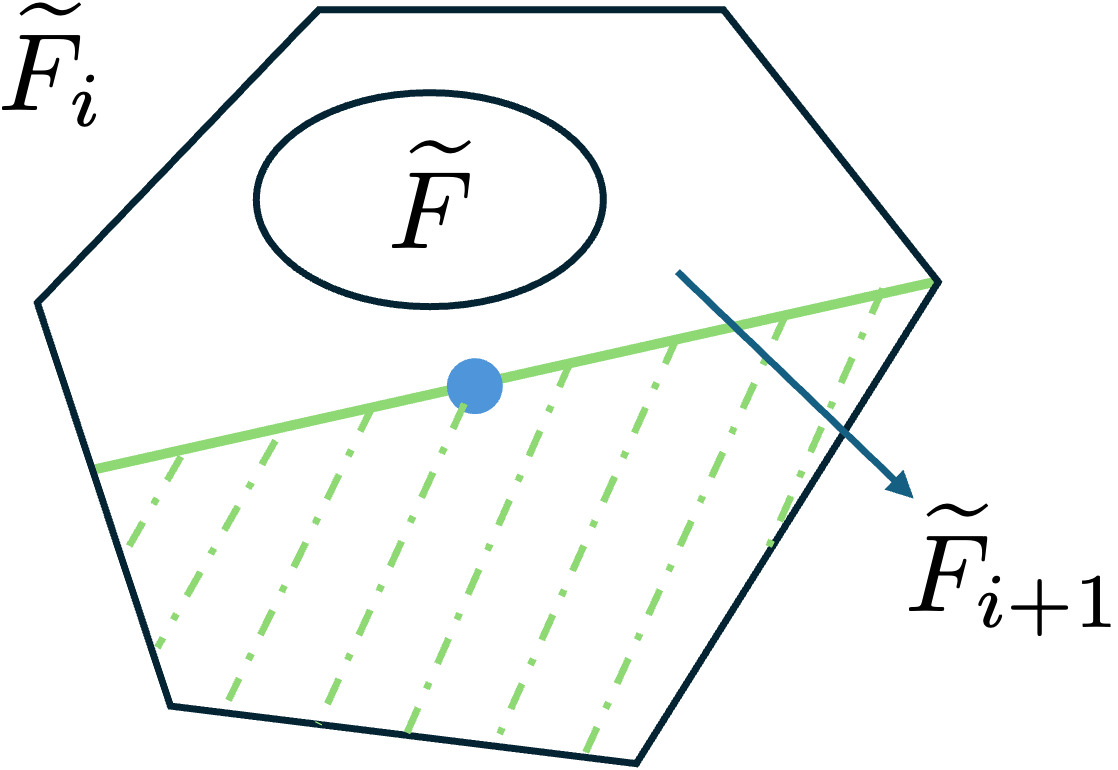}
    \caption{Candidate at the center.}
    \label{fig:cutting_plane_center}
\end{subfigure}
\hfill
\begin{subfigure}{0.49 \columnwidth}
    \centering
    \includegraphics[width= 0.93\textwidth]{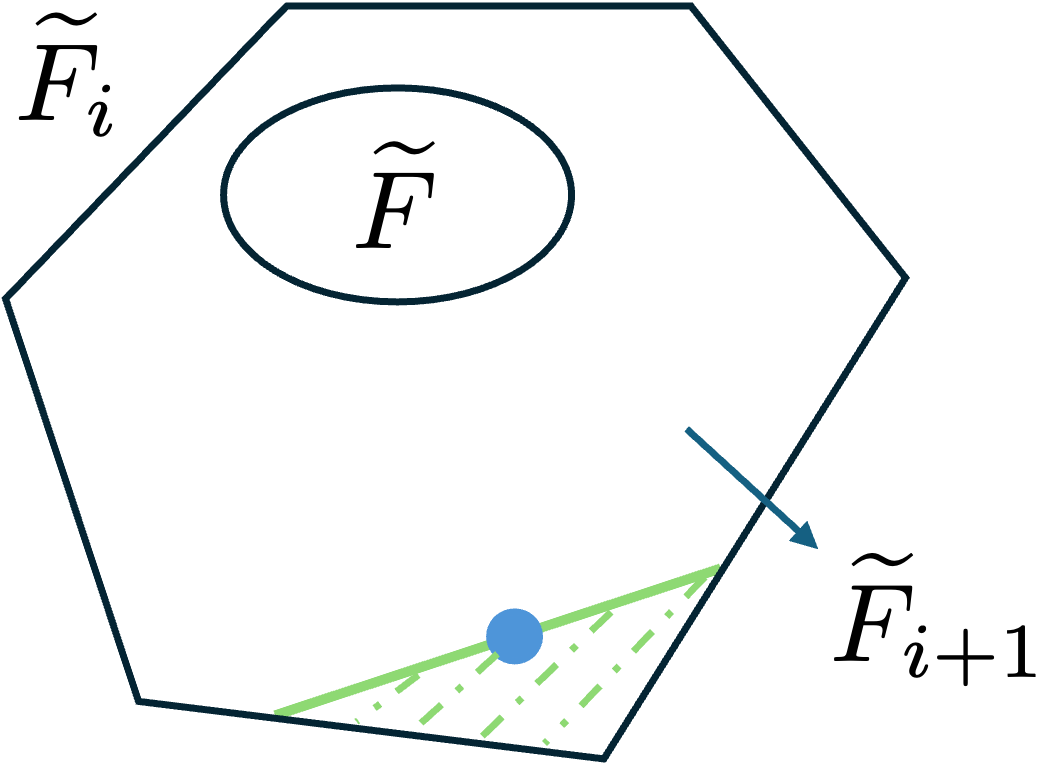}
    \caption{Randomly chosen candidate.}
    \label{fig:cutting_plane_edge}
\end{subfigure}
    \caption{As shown on the left figure, proposing a candidate solution (blue dot) at the center of the localization set is guaranteed to remove a large portion of the search space (shaded area) after verification, while selecting the candidate in an uncontrolled manner (see the right figure) may lead to minimal search space improvement.}
    \label{fig:cutting_plane}
\end{figure}

\subsection{Sampling strategy based on cutting-plane methods}
Cutting-plane methods are iterative algorithms that find a feasible point in a convex target set $\mathcal{F}$ or certify that $\mathcal{F}$ is empty. At each iteration, we maintain an over-approximation $\widetilde{\mathcal{F}} \supseteq \mathcal{F}$ called a localization set. If $\widetilde{\mathcal{F}} = \emptyset$, then we conclude $\mathcal{F} = \emptyset$ and terminate the iteration. Otherwise, we query the `cutting-plane oracle' at a point $W \in \widetilde{\mathcal{F}}$, for which the oracle either certifies that $W \in \mathcal{F}$ or returns a separating hyperplane that separates $W$ and the target set $\mathcal{F}$. In the former case, we terminate the iteration with a feasible $W \in\mathcal{F}$; in the latter case, we update the localization set by adding the separating hyperplane constraint and repeat this process. The intuition behind choosing the candidate solution as the analytic center is illustrated in Fig.~\ref{fig:cutting_plane}.

We design our sampling strategy according to the analytic center cutting-plane method (ACCPM). In the ACCPM, the query point $W$ is chosen as the analytic center~\cite{boyd2004convex} of the localization set $\widetilde{\mathcal{F}}$. Our proposed method consists of a learner, which proposes barrier function candidates based on a set of samples, and a verifier, which serves as a cutting-plane oracle and updates the sample set with counterexamples.


\subsubsection{The learner}
Note that for the candidate barrier function $B(x)$, scaling the parameter $W$ does not affect the satisfaction of conditions~\eqref{eq:cvx_vec_barrier_cond}. Without loss of generality, we search for a feasible parameter $W$ in the bounded set $-R \mathbf{1} \leq W \leq  R \mathbf{1}$ with a constant $R > 0$. Then, for a sample set $\mathcal{S} = \{\mathcal{S}_0, \mathcal{S}_u,  \mathcal{S}_\mathcal{X}\}$ and the corresponding localization set $\widetilde{\mathcal{F}}$, the learner solves the following convex program of finding the analytic center~\cite{boyd2004convex}:
\begin{equation}  \label{eq:ac}
\begin{aligned}
\underset{W}{\text{minimize}} & \ -\sum_{x\in \mathcal{S}_u} \log( B_{i^*}(x)) -\sum_{i=1}^m \sum_{x\in \mathcal{S}_0} \log(-B_i(x)) \\
&\ - \sum_{i = 1}^m \sum_{x \in \mathcal{S}_\mathcal{X}}\log(a_i^\top B(x) - B_i(f_\pi(x)))  \\
& \ + \sum_{i=1}^{(M+1)m}(-\log (W_i + R) - \log (R - W_i) )
\end{aligned}
\end{equation}
where $a_i^\top$ denotes the $i$-th row of $A$ and $W_i$ denotes the $i$-th entry of the vectorized parameter. If the localization set $\widetilde{\mathcal{F}}$ is empty, then we certify that $\mathcal{F}$ is empty. If not, the learner proposes the barrier function candidate $B(x;W^*)$ to the verifier where $W^*$ is the optimal solution to Problem~\eqref{eq:ac}.

\subsubsection{The verifier}
Given a barrier function candidate $B(x;W^*)$ proposed by the learner, the verifier checks if $B(x;W^*)$ satisfies the conditions~\eqref{eq:cvx_vec_barrier_cond} by solving a set of $2m+1$ global optimization problems:
\begin{subequations}
\label{eq:verifier}
\begin{align}
\begin{split}\label{eq:verifier_1}
\underset{x \in \mathcal{X}_u}{\text{min}} \ B_{i^*}(x;W^*),
\end{split} \\
\begin{split}\label{eq:verifier_2}
\underset{x \in \mathcal{X}_0}{\text{min}} \ -B_{i}(x;W^*), 1 \leq i \leq m,
\end{split} \\
\begin{split}\label{eq:verifier_3}
\underset{x \in \mathcal{X}}{\text{min}} \ a_i^\top B(x;W^*) \!-\! B_i(f_\pi(x);W^*), 1 \leq i \leq m.
\end{split}
\end{align}
\end{subequations}

If the optimal values of problems in~\eqref{eq:verifier} are all non-negative, $B(x; W^*)$ is a valid barrier function; otherwise, we extract the optimal solutions $\{x^{ce}_{u, i^*}\}$ of Problem~\eqref{eq:verifier_1}, $\{x^{ce}_{0, i} \}_{i=1}^m$ of Problem~\eqref{eq:verifier_2}, $\{x^{ce}_i\}_{i=1}^m$ of Problem~\eqref{eq:verifier_3} with negative optimal values and denote them as counterexamples~\footnote{For notational simplicity, we let $x^{ce}_{u, i^*}$, $x^{ce}_{0, i}$, or $x^{ce}_i$ be empty if the optimal value of its corresponding problem in~\eqref{eq:verifier} is not negative.}. Then, we expand the sample set $\mathcal{S}$ by
\begin{equation} \label{eq:update_samples}
    \begin{aligned}
        &\mathcal{S}_0 \leftarrow \mathcal{S}_0 \cup \{x^{ce}_{0, i}\}_{i=1}^m, \quad \mathcal{S}_u \leftarrow \mathcal{S}_u \cup \{x^{ce}_{u, i^*}\}, \\
       & \mathcal{S}_\mathcal{X} \leftarrow \mathcal{S}_\mathcal{X} \cup \{x^{ce}_i\}_{i=1}^m.
    \end{aligned}
\end{equation}
The alternation between the learner and the verifier is summarized in Algorithm~\ref{alg:accpm}. Since our CEGIS follows the design of ACCPM, the convergence of Algorithm~\ref{alg:accpm} can be established below. 

\begin{algorithm}[tb]
\caption{ACCPM}
\label{alg:accpm}
\textbf{Input}: Basis function $\phi_\theta(x)$, parameter $A \geq 0$, initial sample set $\mathcal{S}^{(0)} = \{\mathcal{S}_0^{(0)}, \mathcal{S}_u^{(0)}, \mathcal{S}_\mathcal{X}^{(0)}\}$.\\
\textbf{Output}: Parameters $W = \text{vec}(C, b)$ of the NN last linear layer.
\begin{algorithmic}[1] 
\STATE $i = 1$.
\WHILE{True}
\STATE $\mathcal{S}^{(i)} \Rightarrow \widetilde{F}_i$ \algorithmiccomment{See~\eqref{eq:localization}}.
\IF{ $\widetilde{F}_i = \emptyset$}
\STATE Return $W = \emptyset$.
\ENDIF
\STATE Solve \eqref{eq:ac} \COMMENT{Call the learner}.
\STATE Solve~\eqref{eq:verifier} \COMMENT{Call the verifier}.
\IF{$B(x;W_i)$ is a valid barrier function}
\STATE Return $W = W_i$.
\ELSE
\STATE Update the sample set \COMMENT{See~\eqref{eq:update_samples}}.
\ENDIF
\STATE $i=i+1$.
\ENDWHILE
\end{algorithmic}
\end{algorithm}


\begin{theorem} \label{thm:termination}
For the basis function $\phi_\theta(x)$, if the target set $\mathcal{F}$ is non-empty and contains a full dimensional ball of radius $\epsilon>0$ in the parameter space of the last linear layer, Algorithm~\ref{alg:accpm} is guaranteed to find one feasible point in $\mathcal{F}$ in at most $O((m(M+1))^2/\epsilon^2)$ iterations, where $M$ and $m$ denote the input and output dimensions of the last linear layer of $B(x)$ as shown in~\eqref{eq:nn_vbf}.
\end{theorem}

\begin{proof}
Note that the dimension of the parameter space of the last linear layer is $m(M+1)$. The construction of the learner and the verifier follows the analytic center cutting-plane method and the finite-step termination guarantee of Algorithm~\ref{alg:accpm} directly follows from~\cite[Theorem 3.1]{ye1996complexity}.
\end{proof}



\section{Verification-Aided Learning}
\label{sec:training_framework}
We summarize the verification-aided learning framework in Algorithm~\ref{alg:train_verify} that underpins various works~\cite{chang2019neural, abate2021fossil, Dai2021lyapunov, liu2023safe}. As the contributions of this paper, we propose the fine-tuning method in Line $7$, and apply a warm-starting method from~\cite{ash2020warm} in Line 15 to improve the success rate of learning a barrier function. In this section, we explain the implementation details of Algorithm~\ref{alg:train_verify}.

\begin{algorithm}[tb]
\caption{Verification-aided learning}
\label{alg:train_verify}
\textbf{Input}: NN barrier function $B(x)$, initial sample set $\mathcal{S}$, maximum number of iteration $N$, shrinking weight $\lambda \in [0,1]$, noise scale $\sigma \in [0,1].$ \\
\textbf{Output}: Trained NN barrier function $B(x)$, safety certificate.
\begin{algorithmic}[1] 
\FOR{$i = 1, \cdots, N$}
\STATE Train $B(x)$ on $\mathcal{S}$ using gradient descent \COMMENT{See~\eqref{eq:erm}}.
\STATE Verify $B(x)$. \COMMENT{See~\eqref{eq:verifier}}
\IF{Verification succeeds}
\STATE Return $B(x)$ and a safety certificate.
\ELSE
\STATE \textcolor{blue}{Fine-tune the last linear layer by Algorithm~\ref{alg:accpm}.}
\IF{Algorithm~\ref{alg:accpm} is feasible}
\STATE Return tuned $B(x)$ and a safety certificate.
\ENDIF
\STATE Extract counterexamples $\mathcal{S}_{ce}$ for $B(x)$.
\ENDIF
\STATE Augment the counterexample set $\mathcal{S}_{ce}$. 
\STATE Update the sample set $\mathcal{S} \leftarrow \mathcal{S} \cup \mathcal{S}_{ce}$.
\STATE $B(x) = $ Shrink-and-Perturb($B(x)$, $\lambda$, $\sigma$) \COMMENT{See Alg.~\ref{alg:shrink_and_perturb}}.
\ENDFOR
\STATE Return $B(x)$ without any certificate. 
\end{algorithmic}
\end{algorithm}

\begin{algorithm}[tb]
\caption{Shrink-and-Perturb~\cite{ash2020warm}}
\label{alg:shrink_and_perturb}
\textbf{Input}: NN barrier function $B(x)$, shrinking weight $\lambda \in [0,1]$, noise scale $\sigma \in [0,1]$. \\
\textbf{Output}: Adjusted $B(x)$.
    \begin{algorithmic}
        \STATE Randomly initialize a NN $\widetilde{B}(x)$ that has the same architecture as $B(x)$~\footnotemark.
        \FOR{each learnable parameter $W$ in the NN $B(x)$}
        \STATE Select the corresponding parameter $\widetilde{W}$ from $\widetilde{B}$.
        \STATE $W \leftarrow \lambda W + \sigma \widetilde{W}$. 
        \ENDFOR
    \end{algorithmic}
\end{algorithm}

\footnotetext{This step is recommended in~\cite{ash2020warm} since for toolboxes such as PyTorch different network parameters are initialized randomly using different mechanisms.}

\subsection{Train and verify} 
In Line $2$ of Algorithm~\ref{alg:train_verify}, the NN barrier function $B(x)$ is first trained on a sample set by minimizing the empirical risk~\eqref{eq:erm} using gradient descent. Then, the trained $B(x)$ is passed to a complete verifier in Line $3$ which solves a mixed-integer linear program (MILP) for piecewise linear networks~\cite{chen2021learning} or uses satisfiability modulo theory (SMT) solvers such as dReal~\cite{gao2013dreal} for NNs with tanh or sigmoid activation~\cite{peruffo2021automated}. The algorithm terminates immediately when the verification is a success. Otherwise, in Line $7$ we fine-tune the last linear layer of the trained $B(x)$ using Algorithm~\ref{alg:accpm} to further search for a feasible barrier function. 

Since calling the verifier to generate counterexamples can be costly, we first apply PGD to search for counterexamples to falsify the barrier function candidate before doing formal verification in both Line $3$ of Algorithm~\ref{alg:train_verify} and Line $8$ of Algorithm~\ref{alg:accpm}. If the PGD attack finds valid counterexamples, we skip calling the computationally costly verifier and proceed with these counterexamples. 

\subsection{Augment the counterexamples} 
As discussed in~\cite[Section 4.4]{abate2021fossil}, the counterexamples generated by the verifiers are often scarce. Since adding additional samples does not harm the soundness of CEGIS and is in general beneficial for training the NN barrier function, augmenting the counterexamples through sampling has been proposed~\cite{abate2021fossil, Dai2021lyapunov, zhao2022verifying}. In particular, we sample a batch of states around each counterexample in the set $\mathcal{S}_{ce}$ (Line $13$) and use them as the initial points for the PGD attack. The updated states after performing PGD are added to $\mathcal{S}_{ce}$.

\subsection{Warm-starting NN training}
During the training of the barrier function, we want $B(x)$ to not only achieve zero training error shown in~\eqref{eq:erm}, but also zero test error which means the satisfaction of constraints~\eqref{eq:cvx_vec_barrier_cond}. While it is natural to warm-start the barrier function using its parameters learned from the previous iteration in Algorithm~\ref{alg:train_verify}, Ash et al.~\cite{ash2020warm} found that in practice this mechanism often leads to a poorer generalization of $B(x)$ than using fresh random initialization. This implies that warm-starting $B(x)$ from its previous solution can lead to frequent verification failure due to poor generalization. 

Motivated by the fact that the training data gets expanded incrementally in verification-aided learning, we propose to use the shrink-and-perturb trick proposed in~\cite{ash2020warm} to initialize $B(x)$ in each iteration as shown in Algorithm~\ref{alg:shrink_and_perturb}. Shrink-and-perturb is a simple trick to achieve a good balance between generalization improvement and training cost when the training data arrives in a stream and is therefore well-suited for continual learning or active learning tasks. In shrink-and-perturb, the network parameters from the previous training iteration are shrunk toward zero and perturbed by scaled noises from a randomly initialized network. We refer the readers to~\cite{ash2020warm} for a detailed discussion of this trick. 

\subsection{Neural network verifiers}
In our experiments, we primarily focus on the training of ReLU networks which is in fact a piecewise affine function, although our fine-tuning method can be applied to other types of NNs too. To make a comprehensive evaluation of the role of fine-tuning in verification-aided learning, in our experiments, we apply two complete verification methods for ReLU networks. The first method is based on mixed-integer programming~\cite{tjeng2018evaluating} which solves the verification problem by encoding the ReLU network using mixed-integer linear constraints. The second method is based on the GPU-accelerated, specialized NN verifier $\alpha,\beta$-CROWN~\cite{wang2021beta}, which combines efficient linear bounds propagation~\cite{zhang2018efficient, xu2020automatic} and the Branch-and-Bound algorithm for complete verification. $\alpha,\beta$-CROWN is the state-of-the-art NN verifier and has achieved outstanding performance in verifying the robustness of deep NN image classifiers. In this work, we apply it to a control task and deploy it as a complete verifier in Algorithm~\ref{alg:train_verify}. 


\section{Experiments}
\label{sec:simulation}
In this section, we numerically compare the performance of the verification-aided learning framework (Algorithm~\ref{alg:train_verify}) with and without the fine-tuning step on two examples drawn from~\cite{entesari2023reachlipbnb}~\footnote{In our experiments, we used the trained ReLU NN controller from \url{https://github.com/o4lc/ReachLipBnB} for safety verification. }: a double integrator and a 6D quadrotor, both controlled by ReLU networks. We denote Algorithm~\ref{alg:train_verify} with our proposed fine-tuning step (Line $7$) as \texttt{Fine-tuning} and that without Line $7$ as \texttt{Verification-only}. Given the domain $\mathcal{X}$, initial set $\mathcal{X}_0$, and unsafe set $\mathcal{X}_u$, we randomly sample from them to initialize the sample set $\mathcal{S}$. Then, we run Algorithm~\ref{alg:train_verify} multiple times with different random seeds and evaluate the performance of our proposed fine-tuning method by comparing (i) the success rates of finding a valid barrier function and (ii) the runtime of \texttt{Fine-tuning} and \texttt{Verification-only}. In what follows, our experiments show that \texttt{Fine-tuning} can effectively improve both the success rate and runtime compared with \texttt{Verification-only} over multiple runs and for different NN verifiers (see Table~\ref{tab:success_rate_comparison}). In each example, we applied the same set of algorithmic parameters such as the training epochs, the number of augmented counterexample samples, the scaling factors in shrink-and-perturb, etc., to \texttt{Fine-tuning} and \texttt{Verification-only}. The details of these algorithmic parameters can be found in the configuration files in our online code repository \url{https://github.com/ShaoruChen/Neural-Barrier-Function}.


\begin{table}[tb]
    \centering
    \resizebox{\columnwidth}{!}{
    \begin{tabular}{l c c c}
    \toprule
    Example & NN Verifier & \texttt{Fine-tuning} & \texttt{Verification-only} \\ \midrule
    \multirow{2}{0.15\columnwidth}{Double integrator} & MIP & $\mathbf{62.5\%}$ & $37.5\%$ \\ \cmidrule(lr){2-4}
       & $\alpha,\beta$-CROWN & $\mathbf{50\%}$ & $12.5\%$ \\
    \midrule
    \multirow{2}{0.15\columnwidth}{6D quadrotor } & MIP &$\mathbf{100\%}$ & $\mathbf{100\%}$ \\ \cmidrule(lr){2-4} 
    & $\alpha,\beta$-CROWN & $\mathbf{75\%}$ & $12.5\%$ \\
    \bottomrule
    \end{tabular}
    }
    \caption{Success rate comparison of \texttt{Fine-tuning} and \texttt{Verification-only} in learning a valid NN barrier function. For each problem setup, we ran Algorithm~\ref{alg:train_verify} using different random seeds for $8$ times with a timeout threshold of $2$ hours. }
    \label{tab:success_rate_comparison}
\end{table}

\begin{figure}[tb]
\centering
\includegraphics[width=0.95 \columnwidth]{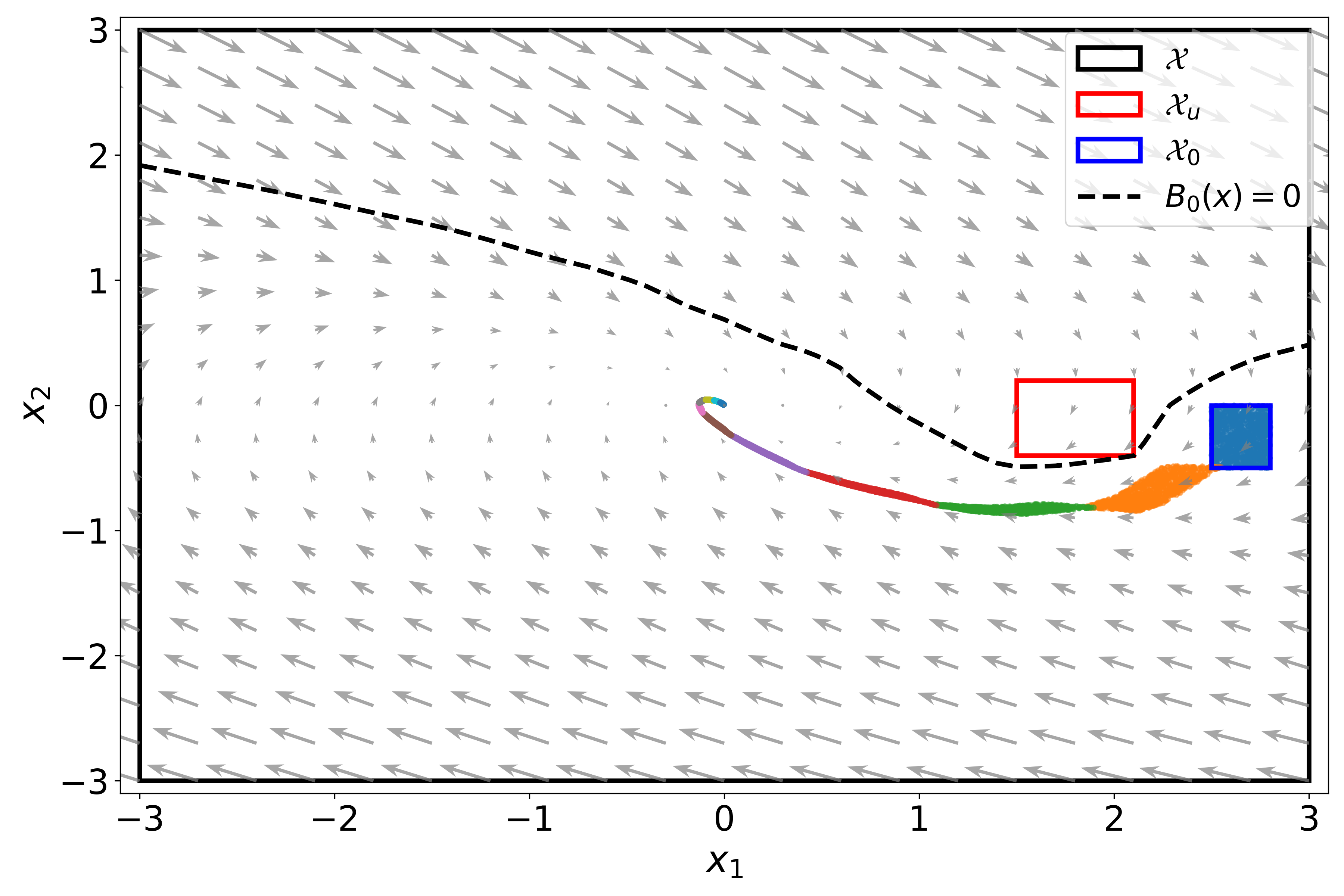}
\caption{The $0$-levelset of $B_0(x)$ (dotted line) separates the trajectory of the double integrator system under a NN controller from the unsafe regions. The gray arrows denote the dynamics of the closed-loop system.}
\label{fig:level_set}
\end{figure}

\subsection{Double integrator}
\begin{figure}
\begin{subfigure}{\columnwidth}
    \centering
    \includegraphics[width=\textwidth]{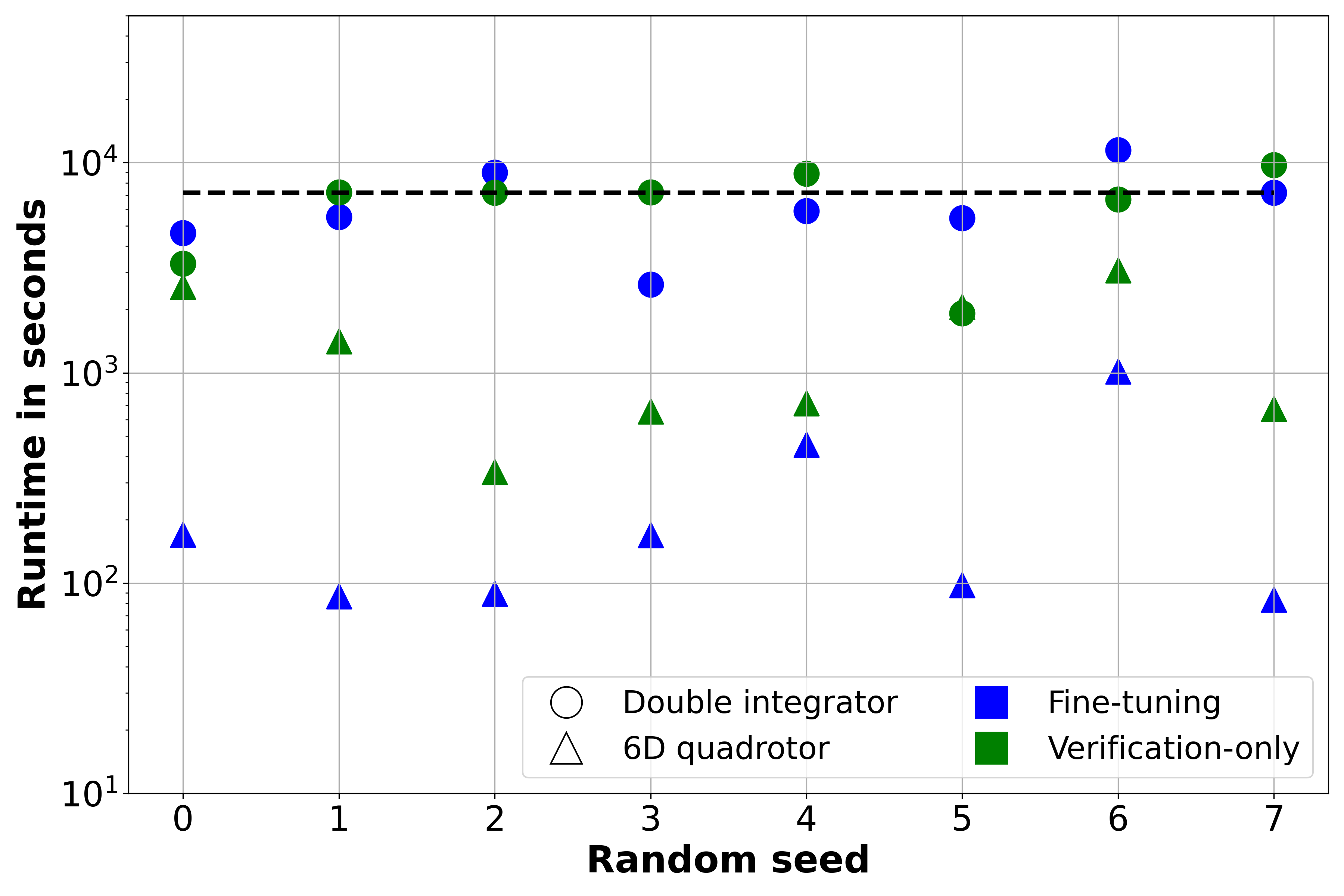}
    \caption{Runtime comparison using MIP-based verifier.}
    \label{fig:runtime_mip}
\end{subfigure}
\begin{subfigure}{\columnwidth}
    \centering
    \includegraphics[width=\textwidth]{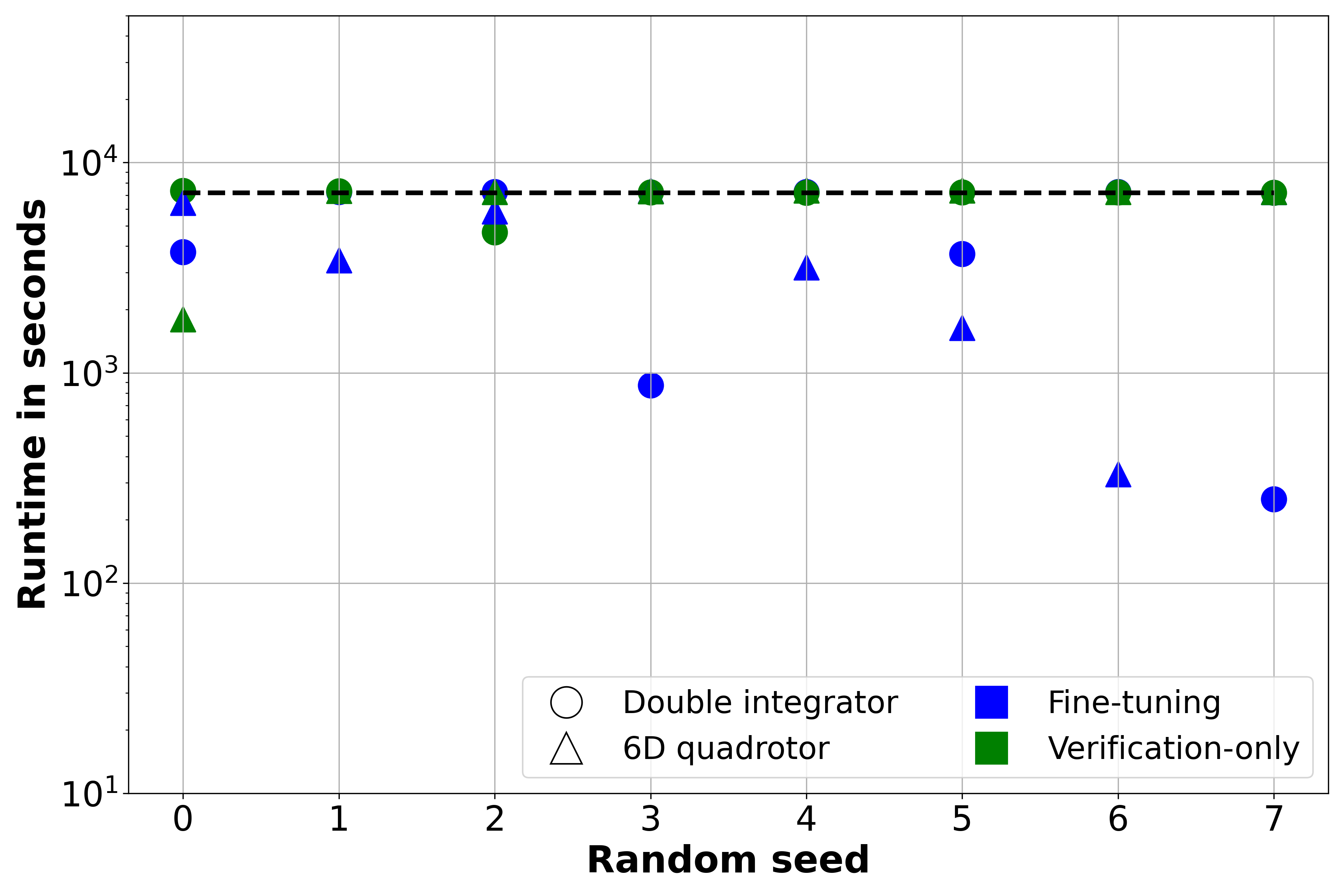}
    \caption{Runtime comparison using $\alpha, \beta$-CROWN.}
    \label{fig:runtime_bab}
\end{subfigure}
\caption{Runtime comparison of Algorithm~\ref{alg:train_verify} with (blue) and without (green) the fine-tuning step for the double integrator (circles) and the 6D quadrotor (triangles) examples. The MIP-based (top) and $\alpha, \beta$-CROWN-based (bottom) NN verifiers are applied, respectively. The dotted line denotes the $2$ hours runtime limit. Any marker below the dotted line denotes a success in finding a valid NN barrier function. }
\label{fig:runtime}
\end{figure}

A discrete-time double integrator system $x_{k+1} = Ax_k + B u_k$ is considered where $A = \bigl[ \begin{smallmatrix}1 & 1\\ 0 & 1\end{smallmatrix}\bigr ]$ and $B= \bigl[ \begin{smallmatrix}0.5 \\ 1 \end{smallmatrix}\bigr ]$. The controller $\pi(x)$ is a ReLU network that has been trained to approximate an MPC policy and has an architecture $2 - 10 -5 -1$. The initial set and unsafe set are given by $\mathcal{X}_0 = \{x \in \mathbb{R}^2 \mid [2.5 \ -0.5]^\top \leq x \leq [2.8 \ 0]^\top  \}$ and $\mathcal{X}_u = \{x \in \mathbb{R}^2 \mid [1.5 \ -0.25]^\top \leq x \leq [1.8 \ 0]^\top  \}$, respectively. The workspace is set as $\mathcal{X}=\{x \in \mathbb{R}^2 \mid [-3 \ -3]^\top \leq x \leq [3 \ 3]^\top \}$ in the example with the MIP-based verifier, and as $\mathcal{X}=\{x \in \mathbb{R}^2 \mid [-0.5 \ -1]^\top \leq x \leq [3 \ 0.5]^\top \}$ when $\alpha, \beta$-CROWN is used~\footnote{While $\alpha, \beta$-CROWN is a complete verifier based on Branch-and-Bound, its current released version is inefficient in finding counterexamples. Therefore, we shrank the workspace for Algorithm~\ref{alg:train_verify} to find feasible barrier functions with $\alpha, \beta$-CROWN as the NN verifier.}. The NN barrier function is parameterized as a ReLU network of architecture $2-30-20-10-5$. We run Algorithm~\ref{alg:train_verify} with a timeout limit of $2$ hours and report the runtime (see the circles) in Fig.~\ref{fig:runtime_mip} and~\ref{fig:runtime_bab} using the MIP-based and $\alpha,\beta$-CROWN NN verifiers, respectively. We observe that \texttt{Fine-tuning} achieves both a higher success rate and a lower average runtime than \texttt{Verification-only}. The validity of a learned NN barrier function is shown in Fig.~\ref{fig:level_set}.

\subsection{6D quadrotor}
Following from~\cite{entesari2023reachlipbnb}, the discretized 6D quadrotor dynamics with sampling time $\Delta t = 0.1$s is given by $x_{k+1} = A x_k + Bu_k + c$ with $A = I_{6 \times 6} + \Delta t \bigl[ \begin{smallmatrix} 0_{3 \times 3} & I_{3\times 3} \\ 0_{3 \times 3} & 0_{3 \times 3} \end{smallmatrix}\bigr ]$, $B = \Delta t \bigl[ \begin{smallmatrix}0 & 0 & 0 & g & 0 & 0\\ 0 & 0 & 0 &  0 & -g & 0 \\ 0 & 0 & 0 &  0 & 0 & 1 \end{smallmatrix}\bigr ]^\top$, and $c = \Delta t \bigl [ \begin{smallmatrix} 0_{5 \times 1}  \\ -g \end{smallmatrix} \bigr ]$. Here $g$ denotes the value of the gravitational acceleration, and $I_{n\times m}$, $0_{n \times m}$ denote the identity and zero matrices of dimension $n \times m$, respectively. The control input $u_k$ is a nonlinear function of the pitch, roll, and thrust of the quadrotor. A ReLU NN controller of architecture $6-32-32-32-3$ is trained to approximate the MPC policy. In this example, the workspace, initial set, and unsafe set are given by $\mathcal{X}=\{x \in \mathbb{R}^6 \mid [4.0 \ 4.0 \ 2.5 \ -1.0 \ -1.0 \ -1.0]^\top \leq x \leq [5.0 \ 5.0 \ 3.5 \ 1.0 \ 1.0 \ 1.0]^\top \}$, $\mathcal{X}_0 =\{x \in \mathbb{R}^6 \mid [4.69 \ 4.65 \  2.975 \  0.9499 \ -0.0001 \ -0.0001]^\top \leq x \leq [4.71 \ 4.75 \ 3.025 \ 0.9501 \ 0.0001 \ 0.0001]^\top \}$, $\mathcal{X}_u=\{x \in \mathbb{R}^6 \mid [4.4  \ 4.3 \ 2.9 \ 0.95 \ -0.1 \ -0.1]^\top \leq x \leq [4.45 \ 4.35 \ 3.0 \ 1.0 \ 0.1 \ 0.1]^\top \}$, respectively. The NN barrier function $B(x)$ is parameterized as a ReLU network of architecture $6-100-80-60-40-20-10$. Similar to the double integrator example, we compare the success rate and runtime of \texttt{Fine-tuning} and \texttt{Verification-only} in Table~\ref{tab:success_rate_comparison} and Fig.~\ref{fig:runtime} and observe that \texttt{Fine-tuning} consistently outperforms \texttt{Verification-only} in both success rate and the average runtime.

\section{Conclusion}
\label{sec:conclusion}
We proposed a NN vector barrier function to certify the safety of NN-controlled systems. To achieve a formal safety guarantee, a verification-aided learning framework is required which iteratively learns the NN barrier function using counterexamples generated by the NN verifier. Motivated by the fact that such "learn-then-verify" iteration may take long to find a valid NN barrier function, we proposed a fine-tuning method for verification-aided learning which enjoys finite-step termination guarantees and demonstrated that our method can effectively improve the success rate and the runtime of synthesizing valid NN barrier functions over numerous experiments. 

\bibliographystyle{IEEEtran}
\bibliography{reference}

\end{document}